%% file: main.tex
\documentclass[11pt,a4paper]{article}
\usepackage[dvipsnames]{xcolor}
\usepackage[hyperref]{emnlp2020}
\usepackage{times}
\usepackage{latexsym}
\usepackage{graphicx}
\usepackage{cleveref}
\usepackage{colortbl}

\usepackage{microtype}

\usepackage{enumitem}
\usepackage{pifont}

\aclfinalcopy 

\renewcommand{\sectionautorefname}{\S\kern-0.2em}
\renewcommand{\subsectionautorefname}{\S\kern-0.2em}
\renewcommand{\subsubsectionautorefname}{\S\kern-0.2em}

\title{Substance over Style: \\
Document-Level Targeted Content Transfer}

\author{Allison Hegel$^1$\thanks{*Work done when the author was at Microsoft Research.} \hspace{1em} Sudha Rao$^2$ \hspace{1em} Asli Celikyilmaz$^2$ \hspace{1em} Bill Dolan$^2$ \\
  $^1$Lexion, Seattle, WA, USA \hspace{1em}
  $^2$Microsoft Research, Redmond, WA, USA \\
  \tt{\small allison@lexion.ai \hspace{2em} \{sudhra,aslicel,billdol\}@microsoft.com} \\
  }

\date{}

\begin{document}
\maketitle
\begin{abstract}
Existing language models excel at writing from scratch, but many real-world scenarios require rewriting an existing document to fit a set of constraints.
Although sentence-level rewriting has been fairly well-studied, little work has addressed the challenge of rewriting an entire document coherently.
In this work, we introduce the task of document-level targeted content transfer and address it in the recipe domain, with a recipe as the document and a dietary restriction (such as \textit{vegan} or \textit{dairy-free}) as the targeted constraint. 
We propose a novel model for this task based on the generative pre-trained language model (GPT-2) and train on a large number of roughly-aligned recipe pairs.\footnote{\url{https://github.com/microsoft/document-level-targeted-content-transfer}}
Both automatic and human evaluations show that our model out-performs existing methods by generating coherent and diverse rewrites that obey the constraint while remaining close to the original document. 
Finally, we analyze our model's rewrites to assess progress toward the goal of making language generation more attuned to constraints that are substantive rather than stylistic.
\end{abstract}

\input{introduction}
\input{dataset}

\input{models}
\input{experiments}
\input{analysis}
\input{related_work}
\input{conclusion}
\input{acknowledgments}

\bibliography{anthology,main}
\bibliographystyle{acl_natbib}

\appendix

\input{appendix}

\end{document}

%% file: introduction.tex
\section{Introduction}

We often think that writing starts from a blank page, but in practice, writing often involves adapting an existing document to fit a new context. This might involve rewriting documentation written for a Mac so that it will apply to a PC, rewriting a lesson plan for a different grade level, or rewriting a product description to appeal to customers in multiple regions.
Automating such rewriting is valuable but challenging, since it requires learning to make coordinated changes spanning an entire document while adhering to constraints that apply not to the style but to the substance of the document.

\begin{figure}[t]
\centering
\includegraphics[scale=0.5]{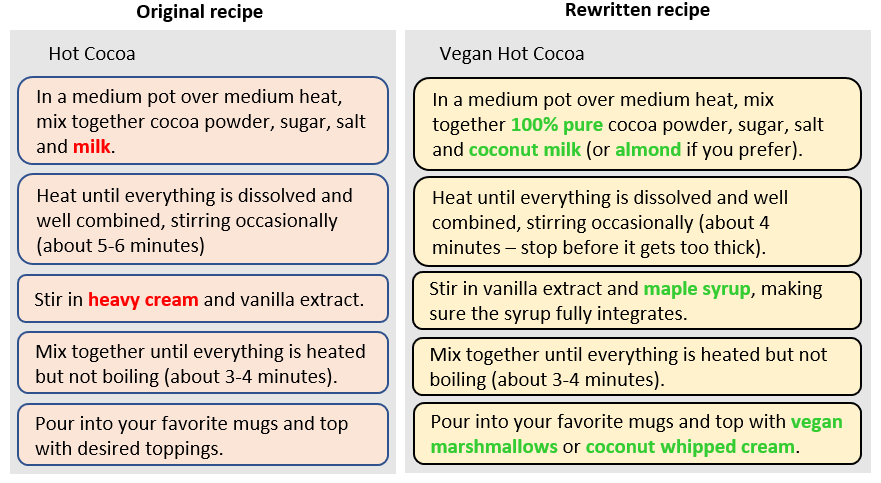}
\caption{\label{fig:recipe-rewite-eg} Document-level targeted content transfer in the recipe domain: given a hot cocoa recipe and the user constraint \textit{vegan}, the task is to rewrite the recipe into a vegan hot cocoa recipe.}
\end{figure}

\begin{figure*}[t]
\centering
\includegraphics[trim=0 130 0 110, scale=0.6]{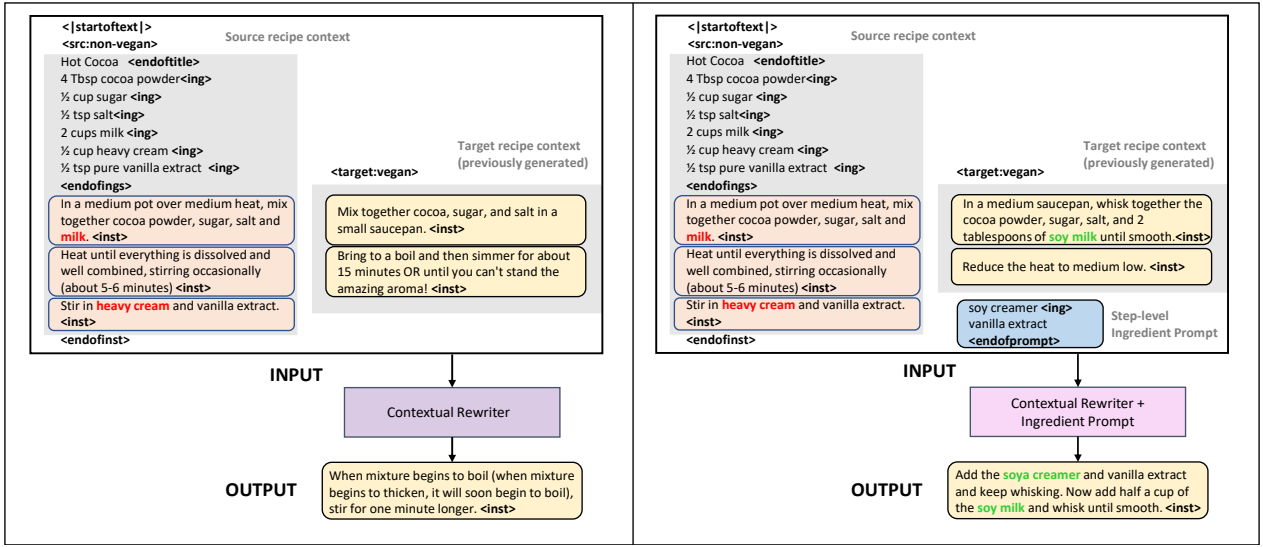}
\caption{\label{fig:model-diagram} Rewrites of the source $n^{th}$ step obtained by the two variants of our proposed model (at test time): (left) Contextual Rewriter, which uses the source context until the $n^{th}$ step and the target context until the $(n-1)^{th}$ step to generate the target $n^{th}$ step; and (right) Contextual Rewriter + Ingredient Prompt, which uses the same context as the previous variant with the addition of a step-level ingredient prompt.}
\end{figure*}

We introduce the novel task of \textbf{document-level targeted content transfer}, defined as rewriting a document to obey a user-provided constraint resulting in some systematic alteration of the document's content.
Success at this task involves both transfer and controlled generation at the document level.
Prior work on controlled generation guides the output of a model using attribute classifiers \citep{Dathathri:19} or control codes \citep{Keskar:19}, but we find that these models do not perform well on our transfer task (\autoref{sec:controllable-baselines}). 
In contrast, models built for the transfer task are generally trained at the sentence level \citep{hu2017toward,Hu:17,Li:18,Rao:18,Syed:19}.

Document-level transfer has typically found success by rewriting each sentence independently \cite{Maruf:19}.
However, many real-world rewriting scenarios require interdependent changes across multiple sentences. 
A clear example is cooking, where rewriting a hot cocoa recipe to make it \textit{vegan} requires more than just substituting ``coconut milk" for ``milk" in a single step—it may also require changing the cooking times and techniques, adjusting ingredient amounts, 
or replacing other ingredients like toppings or spices (\autoref{fig:recipe-rewite-eg}).
Such a rewriting task is substantive rather than stylistic because it changes the content of the recipe, while a stylistic transfer on recipes might instead focus on rewriting a recipe for a different audience, reading level, or writing style such that the content remains the same and only the expression of the recipe changes.

In this work, we address the task of document-level targeted content transfer in the recipe domain, where the document is a recipe and the target constraint is a dietary restriction such as \textit{vegan}.
Given a recipe (source) and a dietary constraint, the task is to rewrite it into a new recipe (target) that obeys the constraint.  
Training a fully-supervised model for this task requires a large number of (recipe, rewritten recipe) pairs, which are difficult to obtain at scale. 
We therefore leverage an alignment algorithm \citep{lin2020recipe} to construct our \textit{noisy} training data pairs
where the source is a recipe that violates a dietary constraint and the target is another recipe for the same dish that obeys the constraint but may not be similar to the source (\autoref{sec:dataset}). 

We propose a novel model for this task which learns to rewrite a source document one step at a time using document-level context.
We start with the recently successful generative pre-trained (GPT-2) language model \citep{Radford:19} and fine-tune it on text that combines \{document-level context, source step, constraint, target step\} using appropriate separators. 
We investigate two variants of our model in the recipe domain:

\paragraph{Contextual Rewriter} (\autoref{sec:context-model}) where the context includes the source recipe (including title, list of ingredients, and steps), any previously rewritten steps, and the targeted constraint (\autoref{fig:model-diagram} left); 

\paragraph{Contextual Rewriter + Ingredient Prompt} (\autoref{sec:ingredient-model}) where, in addition to the context discussed above, we predict a set of step-level ingredients to prompt our rewriter model (\autoref{fig:model-diagram} right).

We compare our proposed models to sentence-level transfer baselines that rewrite each recipe step independently, and to document-level controllable baselines that ignore the source recipe and only control for the dietary constraint (\autoref{sec:baselines-ablations}).
We use automatic metrics and human judgments to evaluate the rewritten recipes, measuring their overall quality, their fluency, their dietary constraint accuracy, and their ability to produce diverse outputs without straying too far from the source recipe (\autoref{sec:eval-metrics}).
Comprehensive experiments demonstrate that our proposed model outperforms baselines by simultaneously accomplishing both transfer and control, but still lacks the substantive knowledge humans rely on to perform well at this task (\autoref{sec:human-rewrite}).
Finally, we conduct an in-depth analysis of various model rewrites and the strengths and weaknesses of the models (\autoref{sec:analysis}).

%% file: dataset.tex
\section{Dataset Creation}
\label{sec:dataset}
The recipe domain, constrained by dietary restrictions, is particularly well-suited to our task since recipes are commonly rewritten according to dietary constraints in real-world scenarios\footnote{In a survey of 250 randomly selected user comments from recipe websites, we found that one third discussed modifying the recipe, often to accommodate dietary restrictions. In addition, U.S. public school cafeterias are required by law to accommodate food allergies and other dietary needs \cite{USDA:17}. Such rewriting that is currently done manually could benefit from our proposed automated approach.}, and this process often requires multiple related changes across the recipe.
To construct our dataset, we use three steps: collect recipes spanning a range of dietary constraints (\autoref{sec:recipes-data}), tag recipes with dietary constraints using a rule-based method (\autoref{sec:apply-constraints}), and align recipes into pairs with similar content but opposite dietary tags (\autoref{sec:align-recipes}). 

Although our model relies on large amounts of parallel data, we obtain this parallel data automatically by running an unsupervised alignment algorithm \cite{lin2020recipe} on non-parallel data. Large collections of non-parallel data are readily available on the web for many other domains, such as lesson plans for different grade levels or technical documentation for different operating systems. With the methods outlined in this section, non-parallel data can be aligned and transformed into a parallel dataset for transfer tasks in other domains. 

\subsection{Collect Recipes}
\label{sec:recipes-data}

We collect English recipes from online recipe websites.\footnote{Websites include Food.com, AllRecipes.com, FoodNetwork.com, and 8 other websites, as well as four existing recipe datasets. Appendix contains full list and associated statistics.}
We remove recipes that lack a title or a list of ingredients, or that have less than two steps.
The resulting dataset contains 1,254,931 recipes, with a median of 9 ingredients and 9 steps.
\subsection{Tag Recipes with Dietary Constraints}
\label{sec:apply-constraints}

We consider seven dietary constraints: dairy-free, nut-free, egg-free, vegan, vegetarian, alcohol-free, and fish-free.\footnote{Each of these constraints is commonly mentioned in recipe titles, and is one of the most common diets \cite{USDA:20} or dietary restrictions \cite{FDA:20}.}
For each dietary constraint, we obtain a list of ingredients that violate it using food lists from Wikipedia.\footnote{E.g. for the dairy-free constraint, we used \url{https://en.wikipedia.org/wiki/Dairy_product}.}
We then compare each recipe's ingredients against that list, and tag it \textit{valid} if there are no violating ingredients, or \textit{invalid} if a violating ingredient is in the recipe.

\subsection{Create Recipe and Step Pairs}
\label{sec:align-recipes}

Our goal is to find recipe pairs for the same dish where one obeys a dietary constraint and the other violates it.
\citet{lin2020recipe} propose a method for automatically aligning two recipes of the same dish. We use their method to first group recipes into dishes, and then find aligned pairs of recipes within a dish where one is valid and the other is invalid.
\autoref{tab:num-recipe-pairs} shows the number of recipe pairs in our dataset for each dietary constraint. It should be noted that these pairs are \textit{noisy} for our rewrite task since the pairs were not created by rewriting. 

The alignment algorithm also gives an alignment score at the step level.
We threshold on this score to keep only the highest-quality step pairs.
Further, in cases where a single source step is aligned to more than one target step with a high score, we combine the target steps together into one, enabling our rewrite model to learn to rewrite one step into multiple steps whenever appropriate. 
\autoref{tab:num-recipe-pairs} (rightmost column) shows the total number of high quality step-level pairs for each dietary constraint that we use to train our rewrite model.

\begin{table}[t]
\centering
\footnotesize
\begin{tabular}{l c c c c}
\hline
 \textbf{Dietary}& \multicolumn{3}{c}{\textbf{Recipe Pairs}} & \textbf{Step Pairs} \\
 \textbf{Constraint}& \textbf{Train} & \textbf{Dev} & \textbf{Test} & \textbf{Train}  \\
\hline
Diary-Free & 194,309 & 10,607 & 9,190 & 2,552,492\\
Nut-Free & 161,596 & 8,722 & 8,989 & 2,060,228 \\
Egg-Free & 124,207 & 5,786 & 5,662 & 1,794,047\\
Vegan & 110,718 & 5,708 & 4,859  & 1,765,865\\
Vegetarian & 59,847 & 2,765 & 2,629  & 682,845\\
Alcohol-Free & 52,157 & 2,348 & 2,136 & 570,627\\
Fish-Free & 34,786 & 1,546 & 1,278 & 383,162 \\
\hline
\end{tabular}
\caption{\label{tab:num-recipe-pairs} Number of recipe pairs and step pairs for each dietary restriction in our data. }
\vspace{-1em}
\end{table}

%% file: models.tex
\section{Model Description}
\label{sec:models}

We propose two model variants for document-level targeted content transfer in the recipe domain. Given a recipe and a dietary constraint, the goal is to rewrite the recipe one step at a time to fit the dietary constraint.

\subsection{Contextual Rewriter}
\label{sec:context-model}

We start with a pre-trained GPT-2 model which is trained on text from 45 million websites with a language modeling objective to predict the next word given previous words.\footnote{This and any future discussion of a pre-trained GPT-2 model refers to the GPT-2 medium model available at \url{https://github.com/huggingface/transformers}.} 
We fine-tune this model using the same language modeling objective on the train split of step-level recipe pairs (\autoref{tab:num-recipe-pairs}).
The left column of \autoref{tab:train-data-format} shows how we format our pairwise data for fine-tuning. 
Given an aligned pair of a source step ($n$) and a target step ($n'$), we prepend the source step $n$ with the source recipe's title, ingredients, and steps from 1 to $(n-1)$; we also prepend the target step $n'$ with target steps from 1 to $(n'-1)$. 
We use separators to demarcate each piece of contextual information. 
Further, to allow the GPT-2 model to understand the dietary constraint, we prepend the entire source-level context with a special tag $<$src:non-constraint$>$ (e.g. non-vegan) and prepend the entire target-level context with a special tag $<$tgt:constraint$>$ (e.g. vegan).

Note that during fine-tuning we use only those steps of a recipe that have been aligned into a pair with a high alignment score (\autoref{sec:align-recipes}). 
However, at test time, we rewrite all steps in the source recipe using the fine-tuned model. 
Also, during fine-tuning, we use the teacher forcing strategy: while rewriting source step $n$, the target recipe context corresponds to the true target steps 1 to $(n'-1)$, whereas during test time, the target recipe context corresponds the previously generated steps 1 to $(n-1)$.\footnote{For decoding, we use top-k sampling $(k=40)$. Appendix contains implementation details for all models.}

\subsection{Contextual Rewriter + Ingredient Prompt}
\label{sec:ingredient-model}

We observe that the rewriter described above often uses ingredients and techniques that diverge from the source recipe. For example, on the left side of \autoref{fig:model-diagram}, the rewritten output diverges from the source recipe when it ignores the ingredients of \textit{``heavy cream and vanilla extract"} in the source step rather than suggesting an appropriate vegan alternative. 
We hypothesize that if the model had the capacity to accept step-level ingredients (in the form of a prompt) as an additional input while rewriting each step, then it could learn to follow the source recipe more closely.
This strategy has proven effective in other domains, including automatic storytelling, where prompting a model with a rough ``storyline" helps models stay on-topic \citep{Yao:18}.

We therefore propose a variant of the previous model that uses step-level ingredients as a prompt in addition to document-level context.
We again start with a pre-trained GPT-2 model and fine-tune it on the train split of step-level recipe pairs (\autoref{tab:num-recipe-pairs}) using a different data format (see the right column of \autoref{tab:train-data-format}).
As in the previous model, we use the source recipe data until step $n$ and the target recipe steps until $(n'-1)$. But before including the target step $n'$, we prompt with the ingredients in $n'$ separated by an $<$ing$>$ separator, and end with an $<$endofprompt$>$ special token. This enables our model to learn to use the ingredient prompt while generating the rewrite. 
\begin{table}[t]
    \centering
    \footnotesize
    \begin{tabular}{l|l}
        $<|$startoftext$|>$ &  $<|$startoftext$|>$\\
        $<$src:non-constraint$>$                     &  $<$src:non-constraint$>$  \\
        src$\_$title $<$endoftitle$>$ & src$\_$title $<$endoftitle$>$  \\
        src$\_$ingredient 1 $<$ing$>$ & src$\_$ingredient 1 $<$ing$>$\\
        ...    &  ... \\
        src$\_$ingredient $K$ & src$\_$ingredient $K$ \\
        $<$endofings$>$                     & $<$endofings$>$ \\
        src$\_$step 1 $<$inst$>$  & src$\_$step 1 $<$inst$>$\\
        ...    & ...\\
        src$\_$step $n$      & src$\_$step $n$ \\
        $<$endofinst$>$                     & $<$endofinst$>$\\
        $<$tgt:constraint$>$                     &  $<$tgt:constraint$>$  \\
        tgt$\_$step 1 $<$inst$>$  & tgt$\_$step 1 $<$inst$>$\\
        ... & ... \\
        tgt$\_$step $n'$  & tgt$\_$step $(n'-1)$  \\
        $<$endofinst$>$ & $<$endofinst$>$ \\
        $<|$endoftext$|>$  & tgt$\_$step $n'$ ingredient 1 $<$ing$>$\\
          & ... \\
          & tgt$\_$step $n'$ ingredient $K_n'$ \\
          & $<$endofprompt$>$ \\
         & tgt$\_$step $n'$\\
            &   $<|$endoftext$|>$\\
    \end{tabular}
    \caption{Data format for fine-tuning a GPT-2 model to rewrite source recipe step $n$ into target recipe step $n'$ (where $n'$ is aligned to $n$) using our Contextual Rewriter (left) and our Contextual Rewriter + Ingredient Prompt (right).}
    \label{tab:train-data-format}
\end{table}

We investigate two methods for generating the step-level ingredient prompt. During fine-tuning, we use the rule-based method. At test time, we generate results using both methods. 

\paragraph{Rule-based ingredient prompt:} Given a source recipe step, we first identify all ingredients mentioned in the step.\footnote{For each ingredient in the recipe's ingredient list, we find the longest n-gram match between ingredient and step, ignoring common recipe stopwords such as ``tablespoons" and descriptors like ``chopped."}
We then use a rule-based method to substitute any ingredients that violate the dietary constraint with alternatives from a food substitution guide \citep{Steen:10}.
While there is work on automatically substituting recipe ingredients with similar ones \citep{Teng:12, Boscarino:14, Yamanishi:15}, to our knowledge no work makes recipe substitutions in accordance with dietary constraints.

\paragraph{GPT-2 ingredient prompt:} We use a GPT-2 model to predict the step-level ingredients to use as prompts. We first collect a dataset of recipe steps from $\sim$1.2 million recipes (from \autoref{sec:recipes-data}). We extract the ingredients from each recipe step using the rule-based method above. We then construct texts by combining \{recipe title, full list of ingredients, steps 1 to $n-1$, ingredients in step $n$\} and fine-tune another GPT-2 model on this text.\footnote{Data format used for fine-tuning is included in appendix.}

%% file: experiments.tex
\section{Experimental Results}
\label{sec:results}
We aim to answer the following research questions:
\begin{enumerate}[noitemsep,nolistsep]
\item Do generation-based rewriters outperform simpler non-learning baselines (\autoref{sec:non-learning-baselines})?  
\item Do our proposed rewriters do a better job of staying close to the source recipe while obeying the constraint compared to controllable generation models (\autoref{sec:controllable-baselines}) that obey the constraint but ignore the source recipe?
\item Do our proposed document-level rewriters outperform sentence-level rewriters (\autoref{sec:sent-level-baselines})? 
\item Does using ingredients as a prompt help our proposed rewriter stay close to the source recipe while obeying the dietary constraint? 
\item Finally, how do models compare to human performance on the rewrite task (\autoref{sec:human-rewrite})?
\end{enumerate}

\subsection{Baselines and Model Ablations}
\label{sec:baselines-ablations}

\subsubsection{Non-learning Baselines}
\label{sec:non-learning-baselines}
\paragraph{Rule-Based:} We use the rule-based method discussed in \autoref{sec:ingredient-model} to rewrite each step independently. This baseline only substitutes ingredients and does not change the cooking times or techniques that may be required for the substitutions to fit.
\paragraph{Retrieval:} We imitate a simple approach to the recipe rewrite task: searching the web for a version of the dish that obeys the given dietary constraint.
Given a source recipe, we determine the dish to which this recipe belongs and retrieve a recipe for the same dish that fits the dietary constraint from the combined pool of train, dev, and test recipes.
\subsubsection{Document-level Controllable Baselines}
\label{sec:controllable-baselines}
We build the following baseline models by providing the title and ingredient list of the target recipe (which obeys the dietary constraint) as the prompt to generate the first target recipe step.
For generating each of the subsequent $n^{th}$ steps, we append the previously generated steps 1 to $(n-1)$ to the prompt. We stop when the model has generated as many steps as there are in the source recipe.
\paragraph{PPLM:} Plug-and-Play Language Model \citep{Dathathri:19} combines a pre-trained language model with a classifier to guide the generation toward a user-specified attribute. We build a PPLM model for our task using a GPT-2 model fine-tuned on $\sim$1.2 million recipes (\autoref{sec:recipes-data}) as the pre-trained language model and using separate bag-of-words classifiers for each of our dietary constraints.\footnote{See appendix for PPLM implementation details.} 
\paragraph{CTRL:} The conditional transformer language model \citep{Keskar:19} uses a `control' code 
to govern the style and content of the generated text. For our task, we use the ``Links" control code to specify the recipe domain.\footnote{See appendix for CTRL implementation details.}

\subsubsection{Sentence-level Transfer Baselines}
\label{sec:sent-level-baselines}
We build additional baseline models for rewriting each step independent of context and train them on our recipe step pairs (\autoref{tab:num-recipe-pairs}). 
\paragraph{Seq2Seq Copy:} We use a sequence-to-sequence model that is enriched with a copy mechanism \citep{jhamtani2017shakespearizing}.
We train separate models for each of our dietary constraints. 
\paragraph{Transformer}
We train a transformer \cite{vaswani2017attention} model with byte-pair encoding.\footnote{We use the implementation at \url{https://github.com/gooppe/transformer-summarization}.}

\begin{table*}[t]
\centering
\footnotesize
\begin{tabular}
{l c | c | c | c }
& \textbf{Fluency} & \textbf{Dietary Constraint} & \textbf{Closeness to Source} & \textbf{Diversity} \\
\textbf{Model} & Perplexity $\downarrow$ & \% Adherence $\uparrow$ & ROUGE $\uparrow$ &  Trigram $\uparrow$\\
\hline
\hline
Non-learning Baselines & & & & \\
\hspace{1.5em} Rule-Based & \multicolumn{1}{r|}{10.24} & 96.1 &  \textit{98.76} & 0.550 \\
\hspace{1.5em} Retrieval & \multicolumn{1}{r|}{\textbf{9.01}} & 93.4 & 28.40 & 0.344 \\
\hline
Document-level Controllable Baselines & & & &\\
\hspace{1.5em} PPLM & \multicolumn{1}{r|}{9.28} & 94.9 & 20.48 & 0.577 \\
\hspace{1.5em} CTRL & \multicolumn{1}{r|}{13.47} & 94.3 & 24.69 & 0.418 \\
\hline
Sentence-level Transfer Baselines & & & & \\
\hspace{1.5em} Seq2seq Copy & \multicolumn{1}{r|}{15.60} & 99.0 & 25.98 & 0.145 \\
\hspace{1.5em} Transformer & \multicolumn{1}{r|}{9.88} & 93.5 & 30.67 & 0.360 \\
\hline
Model Ablations & & & & \\
\hspace{1.5em} No-Source Rewriter & \multicolumn{1}{r|}{\tiny{N/A}} & 96.4 & 20.35 & 0.548 \\
\hspace{1.5em} End-to-End Rewriter & \multicolumn{1}{r|}{9.51} & 97.0 & 25.60 & 0.488 \\
\hspace{1.5em} No-Context Rewriter & \multicolumn{1}{r|}{10.79} &\textbf{99.9} & 31.81 & 0.615 \\
\hspace{1.5em} Contextual Rewriter & \multicolumn{1}{r|}{11.61} & \textbf{99.6} & 31.16 & 0.634 \\
\hspace{1.5em}\hspace{1.5em} + GPT-2 Ingredient Prompt & \multicolumn{1}{r|}{13.86} & \textbf{99.6} & 28.93 & 0.590 \\
\hspace{1.5em}\hspace{1.5em} + Rule Ingredient Prompt & \multicolumn{1}{r|}{12.54} & \textbf{99.5} & \textbf{34.06} & \textbf{0.674} \\
\hline
\end{tabular}
\caption{\label{tab:auto-eval-table} Automatic metric results on model rewrites of 1000 randomly sampled recipes from the test set. The difference between bold and non-bold numbers is statistically significant with $p < 0.001$. We do not compare to \textit{Rule-Based} under closeness to source since it copies steps from the source, leading to an artificially high score.}
\end{table*}

\subsubsection{Model Ablations}

\paragraph{No-Source Rewriter:} We fine-tune a pre-trained GPT-2 model on $\sim$1.2 million recipes (from \autoref{sec:recipes-data}) with a simple language modeling objective. This ablation does not make use of the source recipe, but rather uses only the title and the ingredient list of the aligned target recipe as the prompt, generating the target recipe sequentially.
\paragraph{End-to-End Rewriter:} This model variant is trained end-to-end to rewrite the entire source recipe at once rather than one step at a time. As a prompt, it takes a dietary constraint, a source recipe (title, ingredients and steps), 
and the title and ingredients of the target recipe. We start with a GPT-2 pre-trained model and fine-tune it on the train split of our recipe pair data (\autoref{tab:num-recipe-pairs}) for our task.
\paragraph{No-Context Rewriter:} This variant does not make use of the document-level context, but rather learns to rewrite using only (source step, target step) pairs.
\paragraph{Contextual Rewriter:} This variant makes use of document-level context, but does not use a step-level ingredient prompt.
\paragraph{Contextual Rewriter + GPT-2 Prompt:} At test time, in addition to document-level context, this variant uses the GPT-2 step-level ingredient prediction model (\autoref{sec:ingredient-model}) to generate an ingredient prompt.
\paragraph{Contextual Rewriter + Rule Prompt:} This variant uses the rule-based method (\autoref{sec:ingredient-model}) to generate an ingredient prompt.

\subsection{Evaluation Metrics}
\label{sec:eval-metrics}

\subsubsection{Automatic Metrics}
\label{sec:auto-metrics}

We evaluate model rewrites on 1000 recipes each from the test and dev sets on these criteria:
\paragraph{Fluency:} We measure the perplexity of the model-generated recipes using a GPT-2 language model fine-tuned on recipe data for fair comparison.\footnote{We do not report perplexity for the No-Source Rewriter since we use that model to calculate perplexity.}
\paragraph{Dietary constraint accuracy:} We report the percentage of ingredients in the rewritten recipes that obey the dietary constraint.\footnote{To identify all ingredients in a recipe, we match against a list of foods from \url{https://foodb.ca/}.}
\paragraph{Closeness to source:\footnote{Note that we do not measure closeness to target since we do not have gold target rewritten recipes.}} We report ROUGE-L \cite{lin2002manual} recall score between the source recipe and the rewritten recipe.
\paragraph{Diversity:} Since generation models can produce results that are bland and repetitive, we measure the diversity of the generated recipes in terms of the proportion of unique trigrams \citep{li2015diversitypromoting}.

\subsubsection{Human Judgments}
\label{sec:human-eval}

We conduct human-based evaluation using a crowdsourcing platform\footnote{We use \url{https://www.mturk.com/}. Details on selection, questions, design, and payment in appendix.} on rewrites from the best-performing models based on automatic metrics. 
We randomly sample 150 recipes from our test set with equal proportions of each dietary constraint.
\paragraph{Individual:} We ask 5 judges to rate each rewritten recipe on a scale of 1 to 5 on these criteria:
\paragraph{a. Ingredient usage:} \textit{``Does this recipe use appropriate ingredients for the type of dish it is making?"} 
\paragraph{b. Closeness to source:} \textit{``How close is this recipe to the source while fitting the dietary constraint?''} While some difference from the source is necessary for the rewriting task, this metric evaluates whether the recipe has strayed so far from the source that it may no longer be considered a rewriting of the source recipe.
\paragraph{c. Dietary constraint:} \textit{``Does this recipe fit the specified dietary constraint?"}
\paragraph{d. Overall quality:} \textit{``Is this a good recipe for someone who follows this dietary constraint?"} We expect this metric to indirectly reflect qualities for which there are no well-accepted automatic metrics, such as coherence and the appropriateness of the ingredient prompts.

\paragraph{Comparative:} We also collect human judgments on head-to-head comparisons between models by displaying two rewrites of the same source recipe side by side: one from our best-performing model (Contextual Rewriter + Rule Prompt) and the other from one of the Rule-Based, Retrieval, End-to-End Rewriter, or Contextual Rewriter models. We ask them to choose which of the two rewrites is better overall. Each pairwise comparison is rated by five judges.

\subsection{Automatic Metric Results}

While each model has its strengths, our proposed models provide the best balance of both transfer and control.
\autoref{tab:auto-eval-table} shows the results on model rewrites of 1000 randomly sampled recipes from the test set.\footnote{Results on 1000 recipes from the dev set are reported in the appendix. They follow the same pattern as the test set.}
The retrieval baseline produces the most fluent rewrites, which is expected given that its outputs consist of human-written recipes. However, its scores for closeness to source and adherence to the dietary constraint are considerably lower. 
Document-level controllable baselines produce more diverse outputs than sentence-level transfer baselines, but sentence-level transfer baselines stay closer to the source recipe. 
In particular, Seq2seq Copy achieves a high dietary constraint accuracy, but we noticed that this model generates bland and repetitive outputs (as reflected in its diversity score).
Each of these models has a shortcoming in a key component of the rewrite task.

Under our model ablations, we find that the No-Source Rewriter earns the lowest score for closeness to source, which is predictable given that it does not see the source recipe. 
By introducing source context, the End-to-End Rewriter does slightly better, producing fluent rewrites but still lacking diversity and dietary constraint accuracy.
By rewriting each step independent of context, the No-Context Rewriter achieves a very high dietary constraint accuracy, but does not stay as close to the source as variants that use context. 
The model that introduces a GPT-2 predicted ingredient prompt obeys the dietary constraint well, but is not able to maintain diversity while staying close to the source, suggesting that there is room for improvement in how we build our ingredient prediction model. 
Finally, the rewriter that uses context and a rule-based ingredient prompt performs best across dietary constraint accuracy, closeness to source, and diversity while remaining reasonably fluent.

\begin{table}[t]
\centering
\footnotesize
\begin{tabular}
{l c c c c }
\multicolumn{2}{r}{Ingredient} & Dietary  & Close to & Overall  \\
 Model   &  Usage   & Const. & Source & Quality \\
\hline
\hline
Rule-Based  & 4.64 & 4.70 & \textit{4.58} & 4.47\\
Retrieval  & 4.48 & 4.40 & 3.29 & 3.91\\
End-to-End  & 4.64 & 4.72 & 3.73 & 4.52\\
Contextual & \textbf{4.71} &  \textbf{4.74} & 3.84 & \textbf{4.60}\\
+ Rule Prompt   & 4.67 & \textbf{4.75} & \textbf{4.06} & 4.57\\
\hline
\end{tabular}
\caption{\label{tab:human-eval-table} Human judgments on a scale of 1 to 5 on model rewrites of 150 recipes from test set.}
\vspace{-1em}
\end{table}

\subsection{Human Judgment Results}

\autoref{tab:human-eval-table} shows the results of human judgments on 150 recipe rewrites from the test set.\footnote{Inter-annotator agreement (Krippendorff's alpha) is $0.12$.}
We find that all models except the retrieval baseline achieve similarly high scores.
The Contextual Rewriter + Rule Prompt, the best-performing variant of our model according to automatic metrics, performs well in closeness to source and diversity, reaffirming our previous findings.\footnote{As with automatic metrics, we do not compare to Rule-Based in closeness to source since it copies from the source.} 
Interestingly, the Contextual Rewriter without an ingredient prompt performs better at ingredient usage and receives the highest overall score. Upon further investigation, we find that the rule-based method we used to generate the ingredient prompt sometimes suggests awkward ingredient substitutions such as \textit{``goat soymilk”}, which leads to a lower ingredient usage score. 

\begin{figure}[t]
\includegraphics[scale=0.4]{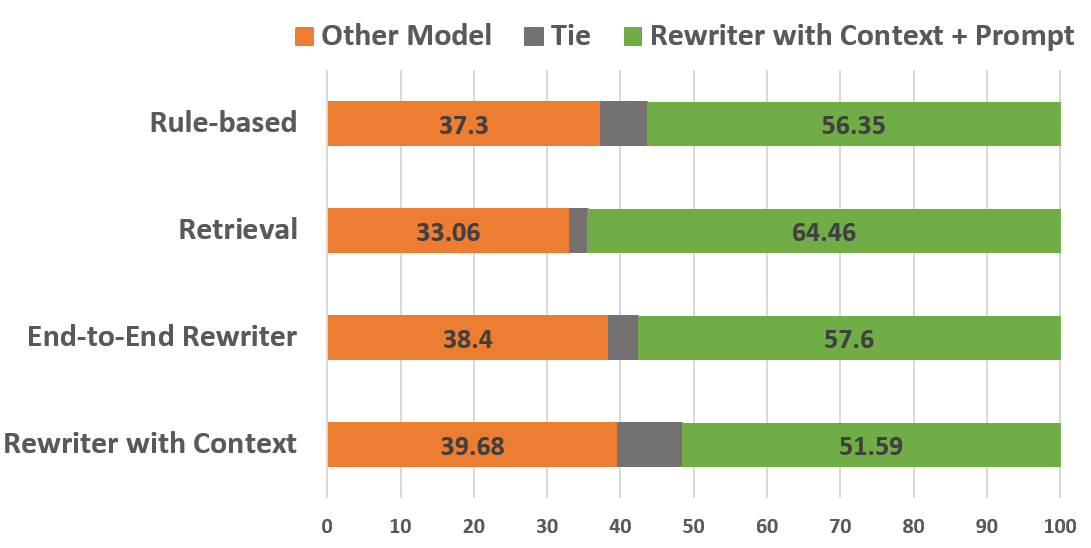}
\caption{\label{fig:human-eval-compare} Results of a pairwise comparison between rewrites of our best model and other models on 150 recipes from the test set as judged by human evaluators. }
\end{figure}

\autoref{fig:human-eval-compare} shows the results of model comparisons.\footnote{Inter-annotator agreement (Krippendorff's alpha) is $0.14$.}
We find that humans prefer our best model considerably over the retrieval baseline, but the Rule-Based method and the End-to-End Rewriter come close to our best model. The Contextual Rewriter performs similarly to our best model.

\subsection{Comparison to Human Rewrite} 
\label{sec:human-rewrite}
We ask three experienced cooks who are current or former vegetarians to rewrite 30 randomly sampled non-vegetarian recipes from our test set into vegetarian recipes.
We find that the human rewrites significantly exceed our best model's performance in all four automatic metrics: fluency (perplexity: 13.91 vs. 20.8), adherence to the dietary constraint (99.7\% vs. 96.3\%), closeness to the source (ROUGE: 77.08 vs. 35.44), and diversity (0.908 vs. 0.836). 
These findings suggest that there is room for further improvement on this task. 

%% file: analysis.tex
\begin{figure*}[t]
\centering
\includegraphics[trim=0 270 0 130, scale=0.58]{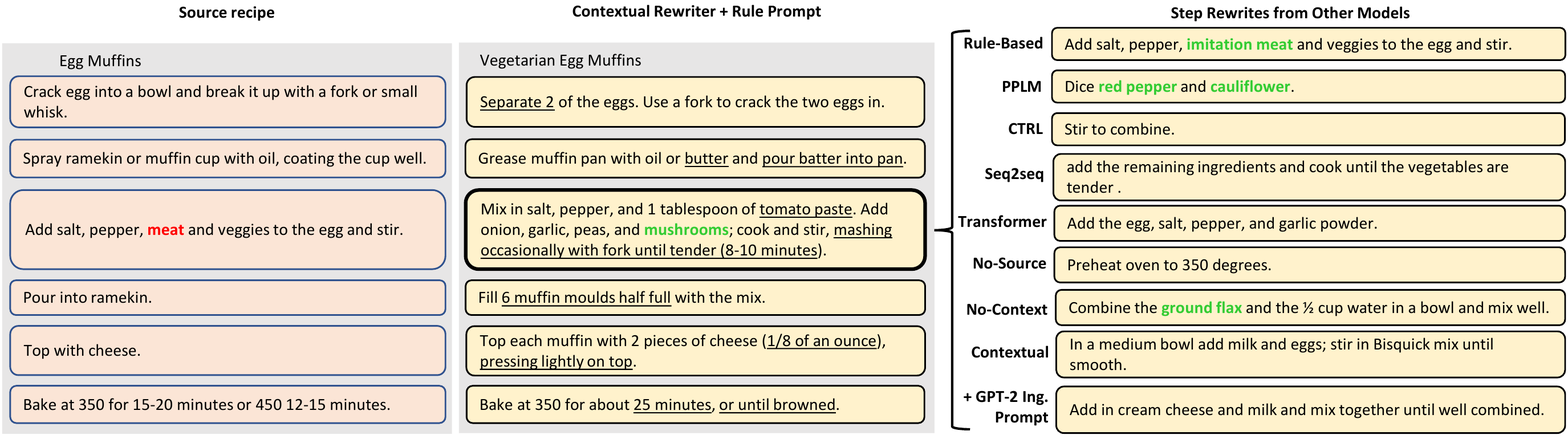}
\caption{\label{fig:rewrite-comparison} A recipe rewritten by the Contextual Rewriter + Rule Prompt, with outputs for a single step from other models for comparison. Our model replaces the violating ingredient (in \textbf{{\color{red}red}}) with a substitution (in \textbf{{\textcolor{Green}{green}}}), as well as modifying or adding new ingredients and techniques in every step (\underline{underlined}).}
\end{figure*}

\section{Analysis}
\label{sec:analysis}

Simple substitution is not adequate for the task of document-level targeted content transfer.
In a recipe that contains a single violating ingredient \textit{``meat"}, the rule-based method makes the minimal edit of substituting \textit{``imitation meat"}, but ignores the other parts of the recipe that must change as a result.
Although on automatic metrics our model does only marginally better, qualitatively we found many cases where the rule-based method fails: it always suggests the same substitutions independent of the type of recipe leading to awkward food combinations, it misses a long tail of uncommon ingredients, and it does not make contextual changes to ingredient amounts, cooking times, or techniques. These flaws lead to the rule-based method performing worse than our model according to human judges (\autoref{tab:human-eval-table} and \autoref{fig:human-eval-compare}).

As \autoref{fig:rewrite-comparison} shows, the Contextual Rewriter + Rule Prompt is capable of more extensive changes based on document-level context. Human evaluators preferred our model's output, which changes multiple ingredients, adds additional techniques, and increases the cooking time. In general, while many of the baseline models tend to produce generic outputs such as \textit{``Preheat the oven"}, our model produces much more diverse recipes and ingredient substitutions.

The larger the number of invalid ingredients for a dietary constraint, the more difficult it was for our model to follow that constraint. Vegan, the most restrictive constraint we studied, had the lowest dietary adherence accuracy across all models (93.6\%). The alcohol-free constraint, which is dominated by one common ingredient (wine), had the highest accuracy (99.5\%) despite the models seeing fewer training examples for that constraint.\footnote{Appendix shows a breakdown of each model's accuracy by dietary restriction.}

The Contextual Rewriter + Rule Prompt falls short in its understanding of the physical entities involved in cooking. Some of the steps it generates are not physically possible, such as \textit{``Dip the cheese into the bread"}.
The model can also suggest unrealistic or illogical cooking times (e.g. \textit{``Bake for 10-10 minutes"}), or change oven temperature mid-recipe. 
While these results are uncommon, they highlight that the model has not learned the physical rules governing the use of ingredients and cooking techniques.

%% file: related_work.tex
\section{Related Work}
\label{sect:related-work}
\paragraph{Text attribute transfer:} 
Most work in text attribute transfer has been at the sentence level, including sentiment \citep{hu2017toward}, formality \citep{Rao:18}, tense \citep{Hu:17}, and authorship \citep{Syed:19}.
While the text attribute transfer approach works well at the sentence level, our work tests a model's ability to make changes across multiple sentences that result in a coherent document.
Further, our method allows for more drastic alterations to the source text than edit-based methods \cite{Li:18} since  we do not restrict the words that the model can alter. 
\paragraph{Document-level controlled generation:} 
The difficulty of text attribute transfer is amplified when the task expands to the document level.
While controlled generation models such as Grover \citep{grover}, PPLM \citep{Dathathri:19}, and CTRL \citep{Keskar:19} have been successful at the document level, they do not make use of context beyond an initial prompt.
For many NLP tasks, contextual information from surrounding sentences can improve the quality of a generated sentence.
We have seen this for coreference resolution \citep{Joshi:19}, relation extraction \citep{Tang:20}, and machine translation \citep{Miculicich:18,Mac:19}. 
In this work, we show the effectiveness of including document-level context when rewriting recipes to fit a dietary constraint. 
\paragraph{Recipe generation:}
Recipe generation has been a research focus for decades, using methods ranging from rule-based planning systems \citep{Hammond:86} to more recent neural network models that use targeted information such as entity types \citep{Parvez:18}, cooking actions \citep{Bosselut:17}, ingredients \citep{Kiddon:16}, or order information \citep{Bosselut:18} to guide the generations. Building on the insight that knowledge about ingredients improves recipe generation, our work uses ingredient prompts to guide the generation of each recipe step. While there has been extensive work on recipe generation, few studies focus on controlled recipe generation. \citet{Majumder:19} recently introduced the task of personalized recipe generation, producing customized recipes based on user preferences. To our knowledge, our work is the first to generate recipes that conform to a given dietary constraint.

%% file: conclusion.tex
\section{Conclusion}
\label{sect:conclusion}

We introduce the novel task of document-level targeted content transfer and address it in the recipe domain, where our documents are recipes and our targeted constraints are dietary restrictions.
We propose a novel model for rewriting a source recipe one step at time by making use of document-level context.
Further, we find that conditioning the model with step-level constraints allows the rewritten recipes to stay closer to the source recipe while successfully obeying the dietary restriction.
We show that our proposed rewriter is able to outperform several existing techniques,
as judged both by automatic metrics and human evaluators. 

Although we focus on the recipe domain, our method naturally generalizes to other domains where procedural tasks can be substantively rewritten. For example, one could rewrite technical documentation by constraining on the target operating system, rewrite lesson plans by constraining on the target grade level, or rewrite furniture assembly instructions by constraining on the tools used.

More broadly, this approach makes it possible to customize existing content to better fit a user's physical reality, whether that entails accommodating their dietary needs, updating their schedule based on the weather forecast, or providing information on a dashboard based on what's in their field of view. As language generation becomes more grounded in signals outside of language, work in the area of substantive transfer becomes increasingly relevant.

%% file: acknowledgments.tex
\section*{Acknowledgments}
\label{sect:acknowledgments}

This work would not have been possible without the support of Microsoft Research's AI Residency program, particularly Becky Tucker and the 2019-20 cohort of residents who maintained a tight-knit research community despite our physical distance.
We would also like to thank Sam Leiboff, Zach Price, and Matt Runchey for their feedback on the recipe content transfer task that informed our modeling approach.
Finally, we are grateful to our anonymous reviewers for their thoughtful comments and suggestions.

%% file: appendix.tex
\section{Dataset Creation}

We collect recipes from recipe websites and existing recipe datasets listed in \autoref{tab:data-sources}.

\begin{table}[htbp]
\centering
\footnotesize
\begin{tabular}{lr}
\hline
\textbf{Recipe Website} & \textbf{Number of Recipes} \\
\hline
AllRecipes.com & 58,535 \\
BBCGoodFood.com & 9,171 \\
Chowhound.com & 3,890 \\
CommonCrawl & 424,621 \\
Epicurious.com \citep{epicurious} & 20,110 \\
Food52.com & 20,595 \\
Food.com & 268,914 \\
FoodNetwork.com & 47,187 \\
Instructables.com & 11,190 \\
MasterCook \citep{nowyourecooking} & 72,141 \\
MealMaster \citep{nowyourecooking} & 312,344 \\
ShowMeTheYummy.com & 555 \\
SimplyRecipes.com & 2,372 \\
SmittenKitchen.com & 986 \\
WikiHow.com & 2,320 \\
\hline
\end{tabular}
\caption{\label{tab:data-sources} Online recipe data sources and amounts.}
\end{table}
While some websites use tags to indicate that a recipe obeys a dietary constraint, not all do, and the tags are often noisy or missing.
We therefore choose not to rely on recipe websites for these tags, and instead we use a rule-based method to tag recipes in our dataset as either \textit{valid} or \textit{invalid} in relation to a dietary constraint. While the method improves our model's performance, we observe several shortcomings. Despite constructing a large set of rules, we still miss words that are uncommon or that did not appear in the train set. Also, since we search for invalid ingredients using the recipe's list of ingredients, we miss ingredients that have been omitted from the ingredient list, as well as ingredients that are not mentioned explicitly by name (e.g. \textit{``fillet"} as in \textit{``catfish fillet"} will not be flagged as an invalid ingredient for a fish-free recipe) or ingredients that are referred to by a brand name or slang term that is not part of our rule set.

While we tried to catch as many of these cases as possible, there are many ambiguous words that the method will incorrectly classify such as \textit{``beefsteak tomato"} appearing to contain meat (\textit{``steak"}), \textit{``oyster crackers"} appearing to contain fish (\textit{``oyster"}), or a variety of \textit{``egg replacer"} brand-name products appearing to contain egg. 

The method is also unable to recognize negation (e.g. \textit{``This recipe is not vegan!"}), or distinguish when a food is marked as optional or as an alternative (e.g. \textit{``Flax is a good substitute for eggs"}). Both of these situations would cause a recipe to be marked with the wrong tag.

After assigning tags, we align similar recipes to form pairs of recipes for the same dish. \autoref{tab:recipe-step-alignment} shows an example alignment between two recipes for Hot Cocoa with the alignment scores for each step. Recipes were divided into 80\% train, 10\% dev, and 10\% test sets before aligning them into pairs, resulting in slightly uneven sizes for each set.

\begin{table*}[htbp]
\centering
\footnotesize
\begin{tabular}{p{0.25cm}p{6.25cm}p{0.25cm}p{6.25cm}p{0.75cm}}
\hline
\textbf{ID} & \textbf{Source Recipe Steps} & \textbf{ID} & \textbf{Target Recipe Steps} & \textbf{Score} \\
\hline
0 & In a medium pot over medium heat, mix together cocoa powder, sugar, salt and milk. & 0 & Heat milk to your desired temperature. & \multicolumn{1}{r}{10.0} \\
 &  & 1 & While milk is being heated, mix hot cocoa mix, creamer, and cinnamon sugar in bowl. & \multicolumn{1}{r}{99.7} \\
 \hline
1 & Heat until everything is dissolved and well combined, stirring occasionally (about 5-6 minutes). & 2 & Add small squirt or about 1/4 teaspoon of chocolate syrup to dry mix. & \multicolumn{1}{r}{1.0} \\
\hline
2 & Stir in heavy cream and vanilla extract. & 3 & Add same amount of syrup again, or enough so that dry mix becomes lumps. & \multicolumn{1}{r}{37.0} \\
\hline
3 & Mix together until everything is heated but not boiling (about 3-4 minutes). & 4 & Add confectioner's sugar and cocoa powder to mix (doesn't have to be as lumpy anymore). & \multicolumn{1}{r}{1.1} \\
\hline
4 & Pour into your favorite mugs and top with desired toppings. & 5 & Pour mix into mug and pour milk on top. & \multicolumn{1}{r}{99.9} \\
 &  & 6 & Add whipped cream and extra chocolate syrup. & \multicolumn{1}{r}{87.4} \\
\hline
\end{tabular}
\caption{\label{tab:recipe-step-alignment} Step-level alignment scores between two Hot Cocoa recipes from the dataset.}
\end{table*}

\section{GPT-2 Model Details}

For each GPT-2 model, we use the 355 million parameter pre-trained GPT-2 medium model. We fine-tune using batch sizes ranging from 2-16 distributed across 64 NVIDIA Tesla V100 GPUs. We use a block size of 1024 for the end-to-end rewriter, and smaller block sizes for models that generate one step at a time of 128 for models without context and 256 for models with context. We train each model for 2 epochs on datasets of aligned recipe steps ranging from 1.4 million to 10 million instances. The No-Context Rewriter was the fastest model to train, at 26 hours per epoch, and the slowest were the End-to-End Rewriter and the Contextual Rewriter + Rule Prompt at 318 hours per epoch.

We experimented with several hyperparameters for generation, including top-k sampling, nucleus sampling, and temperature (\autoref{tab:hyperparameters}) using manually-chosen values. Since most variants performed well in adherence to the dietary constraint, we chose the best-performing variant in perplexity and diversity for our experiments.

\begin{table*}[htbp]
\centering
\footnotesize
\begin{tabular}{l c c c}
\hline
 & Perplexity & \% Adherence & Trigram Diversity \\
\hline
Contextual Rewriter + Rule Prompt & & & \\
\hspace{1.5em} top k = 40, nucleus = 1, temperature = 1 & \textbf{12.54} & 99.5 & \textbf{0.709} \\
\hspace{1.5em} top k = 40, nucleus = 0.8, temperature = 1 & 12.57 & 95.1 & 0.472 \\
\hspace{1.5em} top k = 40, nucleus = 0.9, temperature = 1 & 12.82 & 94.8 & 0.498 \\
\hspace{1.5em} top k = 40, nucleus = 1, temperature = 0.9 & 14.13 & 94.0 & 0.526 \\
\hspace{1.5em} top k = 0, nucleus = 1, temperature = 1 & 17.02 & 99.3 & 0.551 \\
\hspace{1.5em} top k = 10, nucleus = 1, temperature = 1 & 13.98 & 99.3 & 0.492 \\
\hspace{1.5em} top k = 20, nucleus = 1, temperature = 1 & 14.85 & \textbf{99.6} & 0.511 \\
\hline
\end{tabular}
\caption{\label{tab:hyperparameters} Results on the dev set for various generation hyperparameters, including top-k sampling, nucleus sampling, and temperature.}
\end{table*}

We observe that our models can generate diverse rewrites from the same prompt, each with a different degree of fluency and adherence to the dietary constraint. We therefore create a set of rules to select the best generation out of 10 using a set of criteria including use of invalid ingredients, non-dictionary words, and incorrect punctuation.
The criteria for selecting from multiple generations include:
\begin{itemize}[noitemsep,nolistsep]
    \item The step does not contain any violating ingredients
    \item The length is less than 100 characters
    \item The step does not contain special characters including `\%', `*', or `\$'.
    \item The first character is capitalized
    \item The last character is punctuation
    \item All words appear in an English dictionary \citep{pyenchant}
\end{itemize}

\section{Data Format for Document-Level Controllable Baselines}

\paragraph{PPLM}
We use the official codebase for PPLM: \url{https://github.com/uber-research/PPLM}.
To build our PPLM model on our datasets, we use a pre-trained GPT-2 model on $\sim$1.2 million recipes as the pre-trained language model. We build separate bag-of-words classifiers for each of our seven dietary constraints.
We construct the bag-of-words for each dietary constraint by selecting words that appear at least 5 times in recipes fitting the constraint and do not appear in recipes that violate the constraint. At test time, we format the data with the same separators for title, ingredients, and steps used to fine-tune the GPT-2 model on recipe data.

\paragraph{CTRL}
For our task, we use the ``Links" control code to specify the recipe domain. We include the desired dietary restriction in the prompt in addition to the target recipe context and separate them by newlines as they would appear in a web link. We also append the appropriate step number (e.g. ``1.") to the prompt before generating each step.

\section{Data Format for Model Ablations}
We format our recipe data differently for each model ablation described in the main paper. \autoref{tab:ing-pred-data-format} shows the data format we use to fine-tune the GPT-2 model that predicts the ingredients in the next step. \autoref{tab:full-recipe-data-format} shows the data format we use to fine-tune the End-to-End Rewriter. \autoref{tab:simple-multi-data-format} shows the data format we use to fine-tune the No-Context Rewriter. Finally, \autoref{tab:multi-data-format} shows the data format we use to fine-tune the Contextual Rewriter.

\begin{table}[htbp]
    \centering
    \footnotesize
    \begin{tabular}{l}
         $<|$startoftext$|>$ \\
         title $<$endoftitle$>$ \\
         ingredient 1 $<$ing$>$ \\
         ... \\
         ingredient K $<$ing$>$ \\
         $<$endofings$>$ \\
         step 1 $<$inst$>$ \\
         ... \\
         step $(n-1)$ \\
         $<$endofinst$>$ \\
         step $n$ ingredient 1 $<$ing$>$ \\
         ... \\
         step $n$ ingredient $K_n$ \\
         $<|$endoftext$|>$ \\
    \end{tabular}
    \caption{Data format used to fine-tune a GPT-2 model to predict the ingredients in the next step. If there were no ingredients in the next step, we used the token $<$noings$>$.}
    \label{tab:ing-pred-data-format}
\end{table}

\begin{table}[htbp]
    \centering
    \footnotesize
    \begin{tabular}{l}
         $<|$startoftext$|>$ \\
         $<$src:non-constraint$>$ \\
         src$\_$title $<$endoftitle$>$ \\
         src$\_$ingredient 1 $<$ing$>$ \\
         ... \\
         src$\_$ingredient K $<$ing$>$ \\
         $<$endofings$>$ \\
         src$\_$step 1 $<$inst$>$ \\
         ... \\
         src$\_$step $N$ \\
         $<$endofinst$>$ \\
        $<$tgt:constraint$>$ \\
         tgt$\_$title $<$endoftitle$>$ \\
         tgt$\_$ingredient 1 $<$ing$>$ \\
         ... \\
         tgt$\_$ingredient K $<$ing$>$ \\
         $<$endofings$>$ \\
         tgt$\_$step 1 $<$inst$>$ \\
         ... \\
         tgt$\_$step $N$ \\
         $<$endofinst$>$ \\
         $<|$endoftext$|>$ \\
    \end{tabular}
    \caption{Data format used to fine-tune the End-to-End Rewriter.}
    \label{tab:full-recipe-data-format}
\end{table}

\begin{table}[htbp]
    \centering
    \footnotesize
    \begin{tabular}{l}
         $<|$startoftext$|>$ \\
         $<$src:non-constraint$>$ \\
         src$\_$title $<$endoftitle$>$ \\
         src$\_$step $N$ \\
         $<$endofinst$>$ \\
        $<$tgt:constraint$>$ \\
         tgt$\_$step $N$ \\
         $<|$endoftext$|>$ \\
    \end{tabular}
    \caption{Data format used to fine-tune the No-Context Rewriter.}
    \label{tab:simple-multi-data-format}
\end{table}

\begin{table}[htbp]
    \centering
    \footnotesize
    \begin{tabular}{l}
         $<|$startoftext$|>$ \\
         $<$src:non-constraint$>$ \\
         src$\_$title $<$endoftitle$>$ \\
         src$\_$ingredient 1 $<$ing$>$ \\
         ... \\
         src$\_$ingredient K $<$ing$>$ \\
         $<$endofings$>$ \\
         src$\_$step 1 $<$inst$>$ \\
         ... \\
         src$\_$step $(n-1)$ \\
         $<$endofinst$>$ \\
        $<$tgt:constraint$>$ \\
         tgt$\_$step $N$ \\
         $<|$endoftext$|>$ \\
    \end{tabular}
    \caption{Data format used to fine-tune the Contextual Rewriter.}
    \label{tab:multi-data-format}
\end{table}

\section{Example Outputs}

\begin{figure*}[htbp]
\centering
\includegraphics[trim=50 200 0 150, scale=0.67]{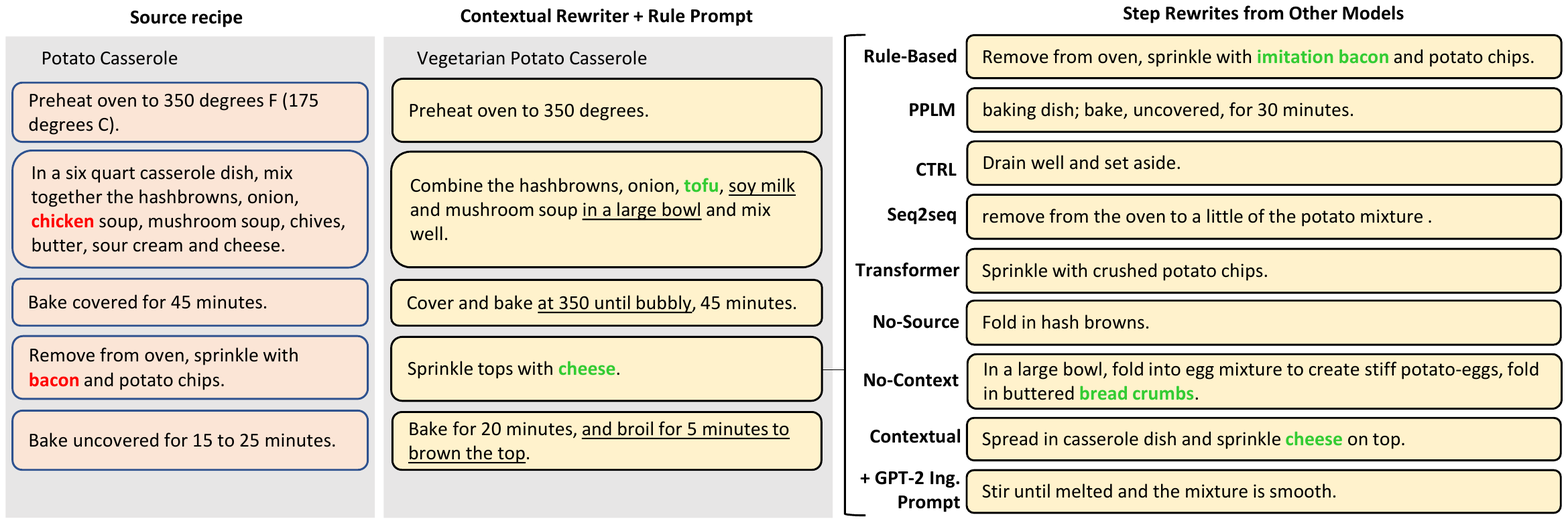}
\caption{\label{fig:rewrite-comparison-appendix} A recipe rewritten by the Contextual Rewriter + Rule Prompt, with outputs for a single step from other models for comparison. Our model replaces the violating ingredient (in \textbf{{\color{red}red}}) with a substitution (in \textbf{{\textcolor{Green}{green}}}), as well as modifying or adding new ingredients and techniques in every step (\underline{underlined}).}
\end{figure*}

\autoref{fig:rewrite-comparison-appendix} shows a source recipe alongside the recipe generated by the Contextual Rewriter + Rule Prompt, as well the generated fourth recipe step from each other model for comparison.

We provide additional step-level examples for each model in \autoref{tab:output-examples-step}, and examples of an entire recipe rewrite for each model in \autoref{tab:output-examples-recipe}. We also show several examples of the ingredient prompts and resulting generations for our two prompt-based models, Contextual Rewriter + GPT-2 Prompt and Contextual Rewriter + Rule Prompt (\autoref{tab:ing-prompt-examples}).

\begin{table*}[htbp]
\centering
\footnotesize
\begin{tabular}{p{5cm}p{10cm}}
\hline
\textbf{Model} & \textbf{Ultimate Green Beans (Vegetarian)} \\
\hline
Original Recipe Step & Cook the {\color{red}bacon} in a large, deep skillet over medium-high heat until crisp, about 10 minutes. \\
\arrayrulecolor{lightgray}\hline
Rule-Based & Cook the {\textcolor{Green}{imitation bacon}} in a large, deep skillet over medium-high heat until crisp, about 10 minutes. \\
\hline
PPLM & Saute {\textcolor{Green}{mushrooms}} in butter until lightly browned. \\
\hline
CTRL & Place the {\textcolor{Green}{mushrooms}}, butter, and sugar in a large pot. \\
\hline
Seq2seq Copy & heat oil in a large pot over medium heat . \\
\hline
Transformer & In a large skillet, heat the oil. \\
\hline
No-Source Rewriter & Saute {\textcolor{Green}{mushrooms}} in butter until soft; set aside. \\
\hline
No-Context Rewriter & WEST FRONT STREET, ST. LOUIS. \\
\hline
Contextual Rewriter & Saute the diced onion in the olive oil over medium heat until golden around the edges. \\
\hline
Contextual Rewriter + GPT-2 Prompt & Chop a bunch of {\textcolor{Green}{mushrooms}}, and slice a small onion. \\
\hline
Contextual Rewriter + Rule Prompt & In large skillet, over medium heat, saute {\textcolor{Green}{mushrooms}} in butter 5 minutes. \\
\hline
Human Rewrite & Peel and sautee garlic and olive oil in large, deep skillet over high heat until slightly scorched skin, about 3 minutes. \\
\arrayrulecolor{black}\hline
 & \textbf{Vegetable Stock (Dairy-Free)} \\
\hline
Original Recipe Step & Melt {\color{red}butter} in 8-quart pan over medium heat. \\
\arrayrulecolor{lightgray}\hline
Rule-Based & Melt {\textcolor{Green}{nondairy butter}} in 8-quart pan over medium heat. \\
\hline
PPLM & Strain through a fine sieve. \\
\hline
CTRL & Add the carrots, celery stalks, green onions, garlic cloves and 1 cup of water. \\
\hline
Seq2seq Copy & heat {\textcolor{Green}{oil}} in a large saucepan over medium heat . \\
\hline
Transformer & In a large saucepan, combine the stock, water, and seasonings. \\
\hline
No-Source Rewriter & Clean and half the vegetables \\
\hline
No-Context Rewriter & Heat the {\textcolor{Green}{olive oil}} in a large pot over medium heat. \\
\hline
Contextual Rewriter & Heat the {\textcolor{Green}{oil}} in a large 4-quart stockpot over medium heat. \\
\hline
Contextual Rewriter + GPT-2 Prompt & Step 4 Then add the trimmings from the carcass and 1 onion and a turnip, and the carrots and celery, and cook until the vegetables are soft, around 4 to 5 hours on a medium heat. \\
\hline
Contextual Rewriter + Rule Prompt & Feel free to play with this amounts; for a cheesy flavour, you can add {\textcolor{Green}{vegan cheese}} or sprinkle {\textcolor{Green}{nutritional yeast}} on top. \\
\arrayrulecolor{black}\hline
 & \textbf{Low Cholesterol Muffins (Egg-Free)} \\
\hline
Original Recipe Step & In another bowl, beat the {\color{red}egg white} and mix together with a fork rapidly for 30 seconds. \\
\arrayrulecolor{lightgray}\hline
Rule-Based & In another bowl, beat the {\textcolor{Green}{egg white substitute}} and mix together with a fork rapidly for 30 seconds. \\
\hline
PPLM & Bake in a preheated oven at 375F for 15 \\
\hline
CTRL & Bake for 20-25 minutes till golden brown. \\
\hline
Seq2seq Copy & in a large bowl , combine the flour , sugar , baking powder , and salt . \\
\hline
Transformer & Bake at 350 degrees for 30 minutes. \\
\hline
No-Source Rewriter & Bring to boil water/juice. \\
\hline
No-Context Rewriter & Mix together dry ingredients. \\
\hline
Contextual Rewriter & In a small bowl, beat {\textcolor{Green}{egg replacer}} with water until fluffy. \\
\hline
Contextual Rewriter + GPT-2 Prompt & Combine 1T globs of the mix onto greased cookie sheets. \\
\hline
Contextual Rewriter + Rule Prompt & Then add {\color{red}egg} or {\textcolor{Green}{egg replacer.}} \\
\arrayrulecolor{black}\hline
 & \textbf{N'Awlins Stuffed Bell Peppers (Fish-Free)} \\
\hline
Original Recipe Step & Add {\color{red}shrimp}, and cook for 2 minutes. \\
\arrayrulecolor{lightgray}\hline
Rule-Based & Add {\textcolor{Green}{tofu}}, and cook for 2 minutes. \\
\hline
PPLM & Makes 8 to 10 servings. \\
\hline
CTRL & To serve, place on serving platter and top each portion with another layer of vegetables. \\
\hline
Seq2seq Copy & add the {\color{red}shrimp} and cook for 5 minutes . \\
\hline
Transformer & Add the {\textcolor{Green}{rice}} and cook, stirring occasionally, until the {\textcolor{Green}{rice}} is tender \\
\hline
No-Source Rewriter & Heat the oven to 350F. \\
\hline
No-Context Rewriter & Simmer covered until the {\textcolor{Green}{peppers}} are tender. \\
\hline
Contextual Rewriter & Add in the minced garlic and red pepper flakes. \\
\hline
Contextual Rewriter + GPT-2 Prompt & Add {\textcolor{Green}{rice}} and simmer just until the liquid has been absorbed. \\
\hline
Contextual Rewriter + Rule Prompt & Add the {\textcolor{Green}{tofu}} and cook for 8 minutes. \\
\arrayrulecolor{black}\hline
\end{tabular}
\caption{\label{tab:output-examples-step} Example step outputs for different dietary constraints. Ingredients that {\color{red}violate} the dietary constraint and their {\textcolor{Green}{substitutions}} are colored.}
\end{table*}

\begin{table*}[htbp]
\centering
\footnotesize
\begin{tabular}{p{2.5cm}p{12.5cm}}
\hline
\textbf{Model} & \textbf{Caramel Popcorn (Dairy-Free)} \\
\hline
Original Recipe & Melt {\color{red}butter}, corn syrup, and brown sugar together. \newline Add {\color{red}sweetened condensed milk} slowly. \newline Cook to soft ball stage. \newline Pour over popcorn and mix. \\
\hline
Rule-Based & Melt {\textcolor{Green}{nondairy butter}}, corn syrup, and brown sugar together. \newline Add sweetened condensed {\textcolor{Green}{soymilk}} slowly. \newline Cook to soft ball stage. \newline Pour over popcorn and mix. \\
\hline
Retrieval & In a large saucepan combine brown sugar, corn syrup, and baking soda. \newline Stir to combine and bring to a soft boil. \newline Add vinegar and boil again. \newline Remove from heat and add vanilla. \newline Place popcorn in a large baking pan. \newline Pour sauce over popcorn and coat evenly. \newline Cover with plastic wrap to keep fresh. \newline Fill brown bags with a scoop of caramel popcorn for take home treats! \\
\hline
PPLM & Combine sugar, {\textcolor{Green}{margarine}}, syrup and salt in a heavy pan. \newline Stir over low heat until mixture is melted and boiling. \newline Boil for 5 minutes, stirring occasionally. \newline Remove from heat. \\
\hline
CTRL & Preheat oven to 350 degrees F. \newline In a large bowl, combine the cornstarch, {\textcolor{Green}{coconut oil}}, and salt. \newline Add the peanuts, sugar, salt, vanilla, and baking soda. \newline Stir until well combined. \\
\hline
Seq2seq Copy & in a large bowl , combine the sugar , brown sugar , and vanilla . \newline add the egg and vanilla extract . \newline cook on low for about 5 minutes . \newline pour into a large bowl and set aside . \\
\hline
Transformer & In a large saucepan, combine brown sugar, corn syrup, and margarine \newline In a large saucepan, combine the brown sugar, corn syrup, and \newline Place popcorn in a large, buttered baking pan. \newline Pour over popcorn and toss to coat. \\
\hline
\raggedright End-to-End Rewriter & Pop popcorn. \newline Remove any un-popped kernels. \newline Put popcorn in a large bowl. \newline Put the nuts on top. \newline In a saucepan, melt {\textcolor{Green}{margarine}}, brown sugar, corn syrup and salt. \newline Bring to a boil, stirring constantly. \newline Boil without stirring for 5 minutes. \newline Remove from heat and add vanilla and soda. \newline Pour syrup over popcorn and mix well. \newline Bake in a preheated 200 degree oven for 1 hour, stirring every 15 minutes. \\
\hline
\raggedright No-Context Rewriter & Combine {\textcolor{Green}{margarine}}, Kahlua, brown sugar, corn syrup and salt in a large, nonstick saucepan. \newline Heat {\textcolor{Green}{margarine}}, brown sugar and corn syrup. \newline Boil for 5 minutes over medium heat without stirring. Stir; bake 5 minutes more. Stir; bake 5 minutes longer. \newline Pour caramel over popcorn and stir well. Spread evenly onto cookie sheet 4 Bake 45 to 55 minutes or until golden brown. Cool completely, about 15 minutes. \\
\hline
\raggedright Contextual Rewriter & In a saucepan, mix brown sugar, {\textcolor{Green}{margarine}}, corn syrup \& salt. \newline Add 2 cups maple syrup, salt, and ground cinnamon to a large saucepan. \newline Cook until the mixture reaches soft ball stage (236 degrees F). \newline Pour the caramel over the popcorn and stir until all of the popcorn is coated. \\
\hline
\raggedright Contextual Rewriter + GPT-2 Prompt & In a large pot, place your popped corn and cover it in the popped corn. \newline Bring to a boil, stirring, then reduce the heat and simmer, stirring once or twice, for 20 minutes or until thick. \newline Continue cooking for 5 minutes while gently stirring once in awhile to stop the edge of the pot from burning. \newline Slowly pour in the corn syrup, and continue mixing until you can form a ball of dough. Roll out dough balls on a board lightly dusted with cornstarch to 1/4 to 1/2-inch thick. \\
\hline
\raggedright Contextual Rewriter + Rule Prompt & Add {\textcolor{Green}{nondairy butter}}, corn syrup and brown sugar to a medium saucepan over medium high heat. \newline Combine the {\textcolor{Green}{soymilk}} and dry ingredients in a medium bowl, then whisk in the wet. \newline Cook over low heat, stirring constantly, about 10 minutes or until thickened; stir twice during cooking. \newline Dump in the popcorn. Stir the mixture to coat it all with corn and pop it in the oven. \\
\hline
\end{tabular}
\caption{\label{tab:output-examples-recipe} Examples of document-level recipe rewrite outputs for Caramel Popcorn when the dietary constraint is dairy-free. Ingredients that {\color{red}violate} the dietary constraint and their {\textcolor{Green}{substitutions}} are colored.}
\end{table*}

\begin{table*}[htbp]
\centering
\footnotesize
\begin{tabular}{p{4cm}p{3cm}p{2cm}p{5cm}}
\hline
\textbf{Source Recipe Step} & \textbf{Model} & \textbf{Ingredient Prompt} & \textbf{Generated Recipe Step} \\
\hline
\raggedright a slice is delicious to eat on its own or with coconut butter, cashew butter, almond butter, or grass fed butter spread on top! & \raggedright Contextual Rewriter + GPT-2 Prompt & \raggedright banana, \newline brown sugar, \newline cinnamon, \newline nutmeg, \newline clove & {\raggedright In a large bowl, stir together banana, brown sugar, egg, butter and nutmeg.} \\
\arrayrulecolor{lightgray}\hline
 & \raggedright Contextual Rewriter + Rule Prompt & \raggedright coconut, \newline sunflower seed butter & {\raggedright Coat with your favorite healthy mix of seeds butter a few drops of liquid stevia or sweetener of choice sprinkled on top (optional).} \\
\arrayrulecolor{black}\hline
\raggedright Serve with milk or with a lump of butter melting into the nice warm pudding. & \raggedright Contextual Rewriter + GPT-2 Prompt & \raggedright spray oil, \newline vanilla, \newline chocolate bread pudding & {\raggedright Spray with cooking spray and add the chocolate bread pudding.} \\
\arrayrulecolor{lightgray}\hline
 & \raggedright Contextual Rewriter + Rule Prompt & \raggedright soymilk, \newline nondairy butter, \newline pudding & {\raggedright Top pudding with dollops of vegan butter and a sprinkle of cinnamon.} \\
\arrayrulecolor{black}\hline
\raggedright Stir in beer. & \raggedright Contextual Rewriter + GPT-2 Prompt & \raggedright coconut, \newline paprika & {\raggedright Heat up about 2 to 3 inches of coconut oil in a cast iron skillet.} \\
\arrayrulecolor{lightgray}\hline
 & \raggedright Contextual Rewriter + Rule Prompt & \raggedright non-alcoholic beer & {\raggedright Slowly pour in 3/4 cup non-alcoholic beer, whisking briskly to blend.} \\
\arrayrulecolor{black}\hline
\end{tabular}
\caption{\label{tab:ing-prompt-examples} Examples to show the effect of the predicted and rule-based ingredient prompts on the generated steps.}
\end{table*}

\section{Additional Results}

We provide the automatic metric results for 1000 recipes randomly sampled from the dev set in \autoref{tab:tune-auto-eval-table}. We also provide a detailed breakdown of each model's accuracy across the seven dietary constraints in \autoref{tab:dietary-accuracy-breakdown}. Finally, we show a comparison of the results for human-written recipe rewrites against our best model, the Contextual Rewriter + Rule Prompt, on a subset of 30 vegetarian recipes from the test set (\autoref{tab:human-vs-machine}).

\begin{table*}[t]
\centering
\footnotesize
\begin{tabular}
{l c | c | c | c }
& \textbf{Fluency} & \textbf{Dietary Const.} & \textbf{Closeness to Source} & \textbf{Diversity} \\
\textbf{Model} & Perplexity $\downarrow$ & \% Adherence $\uparrow$ & ROUGE $\uparrow$ &  Trigram $\uparrow$\\
\hline
Non-learning & & & & \\
\hspace{1.5em} Rule-Based & \multicolumn{1}{r|}{10.92} & 96.6 & \textit{98.77} &  0.557 \\
\hspace{1.5em} Retrieval & \multicolumn{1}{r|}{11.09} & 93.9 &  26.78 &  0.380 \\
\hline
Controllable Generation & & & &\\
\hspace{1.5em} GPT-2 & \multicolumn{1}{r|}{\tiny{N/A}} & 97.6 & 21.14 &  0.530 \\
\hspace{1.5em} PPLM & \multicolumn{1}{r|}{12.85} & 95.2 & 20.83 &  0.577 \\
\hspace{1.5em} CTRL & \multicolumn{1}{r|}{13.14} & 94.6 & 25.52 &  0.433 \\
\hline
Sentence-level Transfer & & & & \\
\hspace{1.5em} Seq2seq Copy & \multicolumn{1}{r|}{16.72} & 98.6 &  25.80 & 0.144 \\
\hspace{1.5em} Transformer & \multicolumn{1}{r|}{9.85} & 96.3 &  27.54 & 0.328  \\
\hline
Proposed Model Ablations & & & & \\
\hspace{1.5em} End-to-End Rewriter & \multicolumn{1}{r|}{\textbf{9.69}} & 97.6 & 25.81 &  0.481 \\
\hspace{1.5em} No-Context Rewriter & \multicolumn{1}{r|}{13.91} & \textbf{99.9} & 32.13 & 0.591 \\
\hspace{1.5em} Contextual Rewriter & \multicolumn{1}{r|}{12.38} & 97.1  & 31.28 &  0.652 \\
\hspace{1.5em}\hspace{1.5em} + GPT-2 Ingredient Prompt & \multicolumn{1}{r|}{16.37} & \textbf{99.8} & 29.36 & 0.573 \\
\hspace{1.5em}\hspace{1.5em} + Rule Ingredient Prompt & \multicolumn{1}{r|}{14.60} & \textbf{99.8}  & \textbf{35.08} &  \textbf{0.709} \\
\hline
\end{tabular}
\caption{\label{tab:tune-auto-eval-table} Automatic metric results on model rewrites of 1000 randomly sampled recipes from the dev set. The difference between bold and non-bold numbers is statistically significant with $p < 0.001$. We do not compare to \textit{rule-based} under closeness to source since it copies steps from the source, leading to an artificially high score. }
\end{table*}

\begin{table*}[t]
\centering
\footnotesize
\begin{tabular}
{l r | r r r r r r r}
\textbf{Model} & \textbf{Overall} & \textbf{Dairy} & \textbf{Nut-Free} & \textbf{Egg-Free} & \textbf{Vegan} & \textbf{Veget.} & \textbf{Alc.-Free} & \textbf{Fish-Free} \\
\hline
Non-learning & & & & & \\
\hspace{1.5em} Rule-Based & 96.1 & 95.1 & 96.9 & 96.5 & 93.5 & 98.6 & 98.9 & 97.8 \\
\hspace{1.5em} Retrieval & 93.4 & 91.9 & 99.2 & 95.5 & 84.9 & 92.8 & 96.4 & 98.8 \\
\hline
Controllable Generation & & & & & \\
\hspace{1.5em} GPT-2 & 96.4 & 95.9 & 98.5 & 99.3 & 91.1 & 96.0 & 99.8 & 100.0 \\
\hspace{1.5em} PPLM & 94.9 & 92.9 & 97.6 & 99.5 & 89.1 & 93.6 & 100.0 & 100.0 \\
\hspace{1.5em} CTRL & 94.3 & 92.3 & 95.8 & 95.6 & 90.1 & 95.4 & 100.0 & 100.0 \\
\hline
Sentence-level Transfer & & & & & \\
\hspace{1.5em} Seq2seq Copy & 99.0 & 97.2 & 100.0 & 100.0 & 99.3 & 99.1 & 100.0 & 99.3 \\
\hspace{1.5em} Transformer & 93.5 & 89.8 & 98.1 & 98.7 & 87.5 & 92.2 & 98.7 & 100.0 \\
\hline
Proposed Model Ablations & & & & & \\
\hspace{1.5em} End-to-End Rewriter & 97.0 & 97.1 & 99.4 & 98.4 & 91.8 & 96.1 & 100.0 & 100.0 \\
\hspace{1.5em} No-Context Rewriter & 99.9 & 100.0 & 100.0 & 100 & 99.8 & 100.0 & 100.0 & 100.0 \\
\hspace{1.5em} Contextual Rewriter & 99.6 & 99.9 & 100.0 & 100.0 & 98.5 & 99.1 & 100.0 & 100.0 \\
\hspace{1.5em} + GPT-2 Ing. Prompt & 99.6 & 99.7 & 99.7 & 99.6 & 98.9 & 99.5 & 100.0 & 100.0 \\
\hspace{1.5em} + Rule Ing. Prompt & 99.5 & 99.7 & 99.7 & 100 & 99.2 & 98.2 & 100.0 & 99.2 \\
\hline
\end{tabular}
\caption{\label{tab:dietary-accuracy-breakdown} Further detail on dietary constraint accuracy for 1000 randomly sampled recipes from the test set.}
\end{table*}

\begin{table*}[t]
\centering
\footnotesize
\begin{tabular}
{l c | c | c | c }
& \textbf{Fluency} & \textbf{Dietary Const.} & \textbf{Closeness to Source} & \textbf{Diversity} \\
\textbf{Model} & Perplexity $\downarrow$ & \% Adherence $\uparrow$ & ROUGE $\uparrow$ &  Trigram $\uparrow$\\
\hline
Human Rewrite & \multicolumn{1}{r|}{13.91} & 99.7 & 77.08 & 0.906 \\
Contextual Rewriter + Rule Ing. Prompt & \multicolumn{1}{r|}{20.28} & 96.3  & 35.44 &  0.836 \\
\hline
\end{tabular}
\caption{\label{tab:human-vs-machine} Comparison of the rewrites done by humans to the Contextual Rewriter + Rule Prompt on a subset of 30 vegetarian recipes from the test set. }
\end{table*}

\section{Human Evaluation}

For human evaluation, we limited our annotators to workers who met the following criteria:
\begin{itemize}[noitemsep,nolistsep]
    \item HIT Approval Rate (\%) for all Requesters' HITs greater than 90
    \item Location is one of AU, CA, NZ, GB, US
    \item Number of HITs Approved greater than 500
    \item Masters has been granted (user was identified by the platform as a high-performing annotator)
\end{itemize}

\begin{figure*}[htbp]
\centering
\includegraphics[trim=0 0 0 0, scale=1]{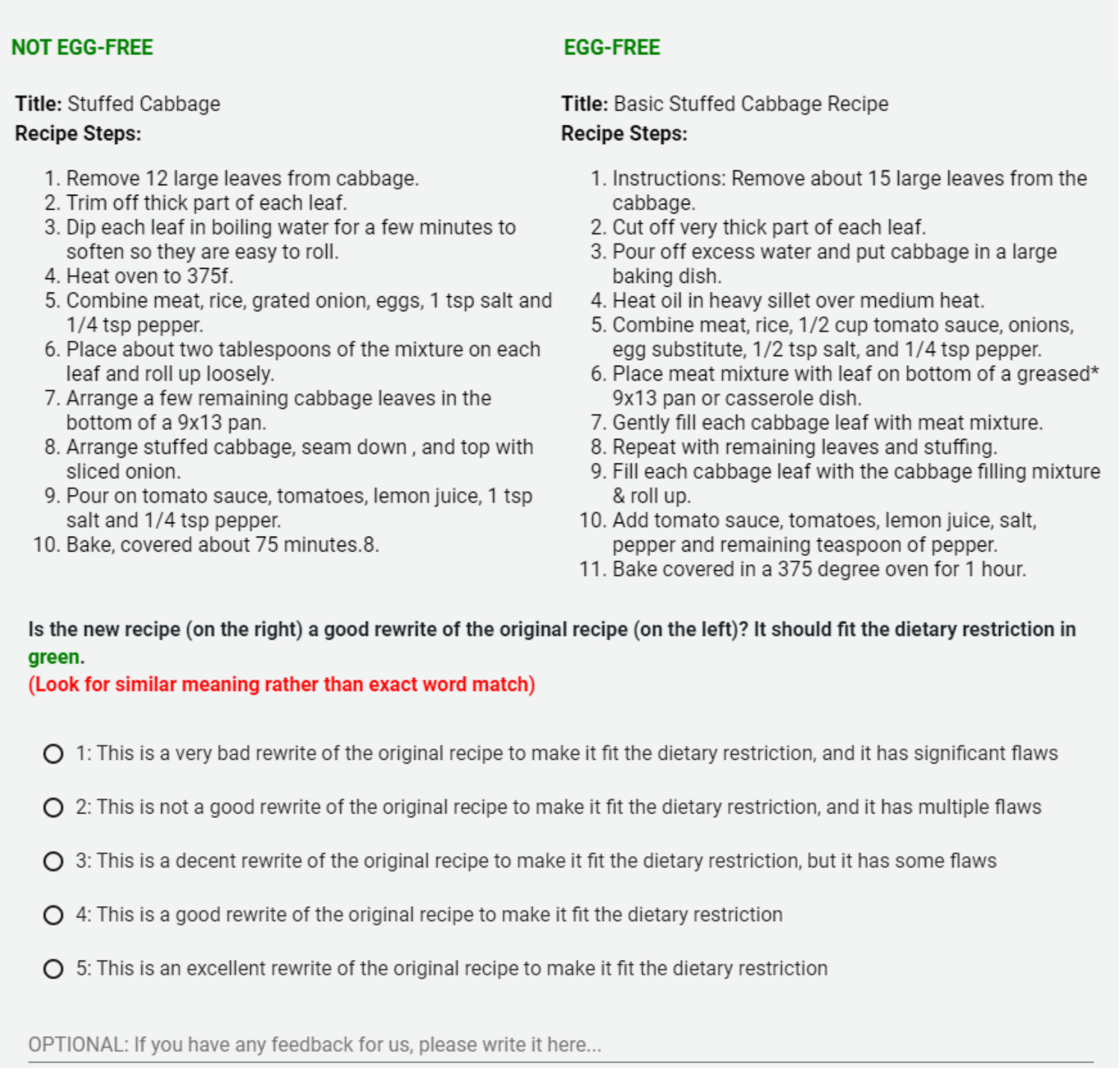}
\caption{\label{fig:mturk1} An example of a question used for human evaluation of the recipe rewrite task on Amazon Mechanical Turk.}
\end{figure*}

\begin{figure*}[htbp]
\centering
\includegraphics[trim=0 0 0 0, scale=1]{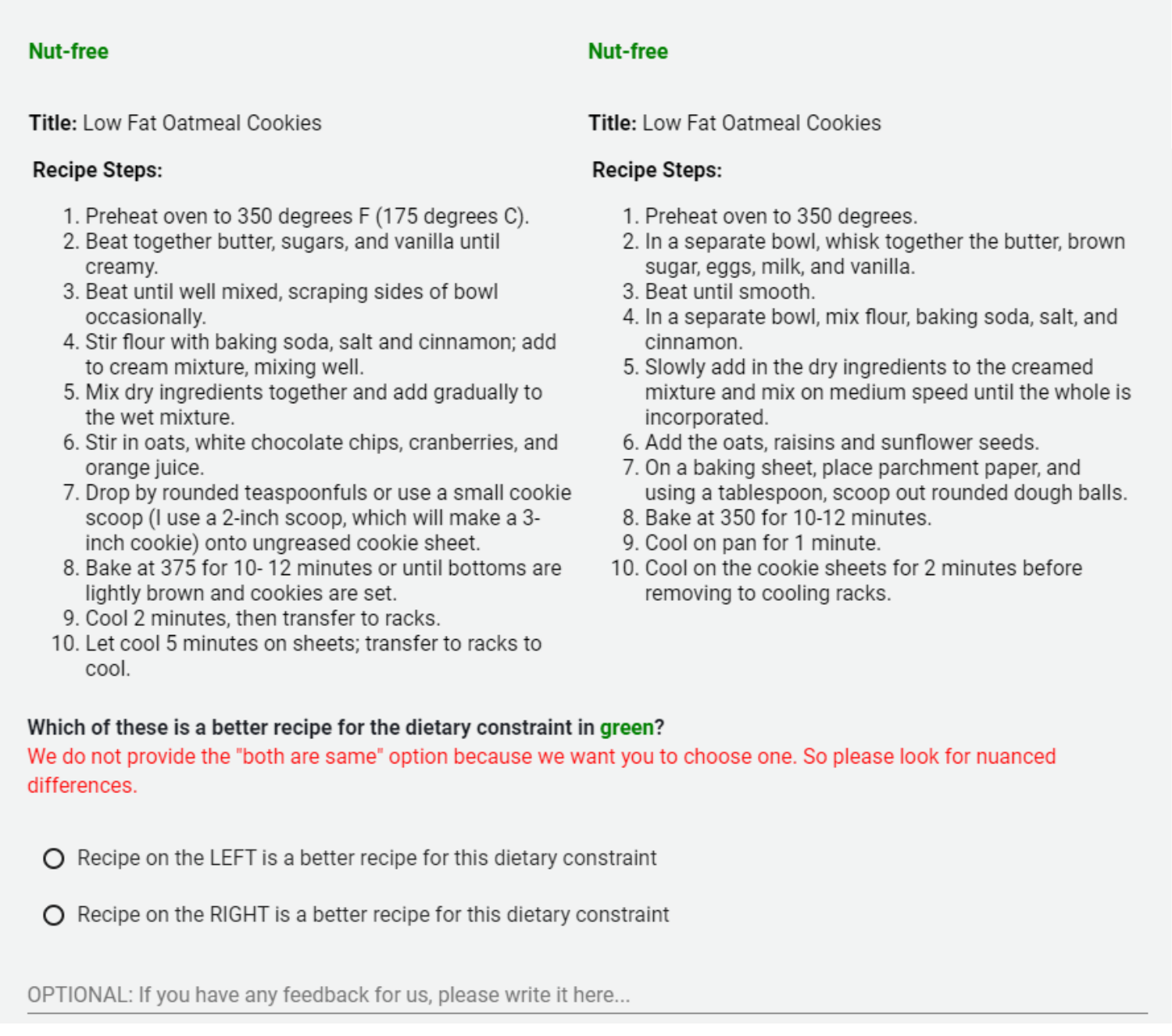}
\caption{\label{fig:mturk2} An example of a question used for human evaluation of the recipe rewrite task on Amazon Mechanical Turk.}
\end{figure*}

\begin{figure*}[htbp]
\centering
\includegraphics[trim=0 0 0 0, scale=0.8]{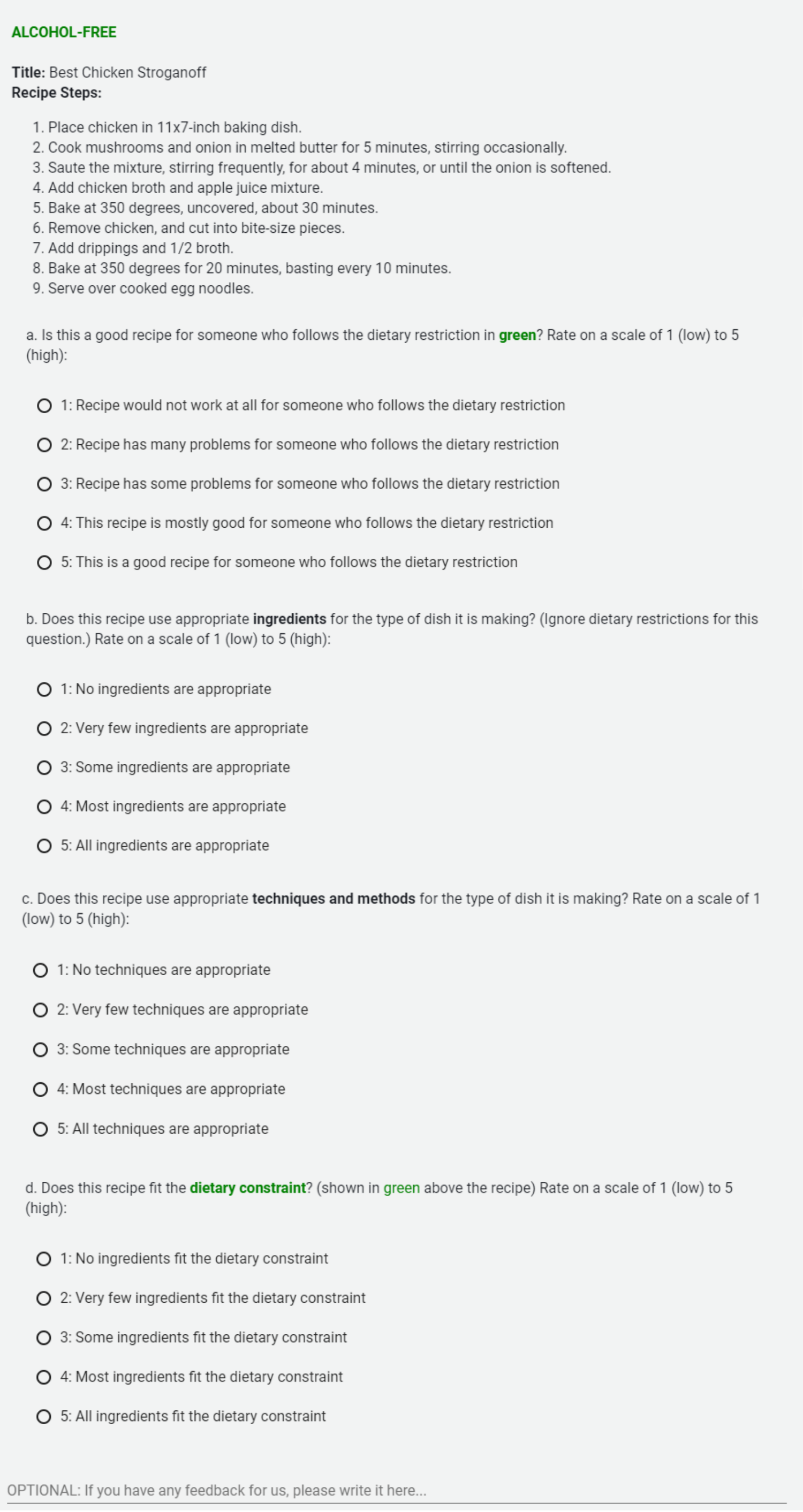}
\caption{\label{fig:mturk3} An example of a question used for human evaluation of the recipe rewrite task on Amazon Mechanical Turk.}
\end{figure*}

We obtained 5 evaluations per recipe for each of the questions listed in \autoref{fig:mturk1} (paying \$0.30 per response), \autoref{fig:mturk2} (\$0.25), and \autoref{fig:mturk3} (\$0.50). For the head-to-head model comparison, if fewer than 3 of the 5 evaluations agreed, we considered it a tie between the models. We did not have our human annotators evaluate the fish-free dietary constraint since the most common violating ingredient, Worcestershire sauce, is not commonly known to contain fish, which caused our annotators confusion in an initial test run.

